\documentclass[letterpaper, 10 pt, conference]{ieeeconf}  

\IEEEoverridecommandlockouts                              

\usepackage{kotex}
\usepackage{cite}
\usepackage{amsmath}
\usepackage{amsfonts}
\let\labelindent\relax
\usepackage[inline]{enumitem}
\usepackage{graphicx}
\usepackage{caption}
\usepackage{subcaption}
\usepackage{lipsum}
\usepackage{algorithm}
\usepackage{algorithmicx}
\usepackage{algpseudocode}
\usepackage{makecell}
\usepackage{verbatim}
\usepackage{multirow}
\usepackage{multicol}
\usepackage{flushend}
\usepackage{xcolor}

\usepackage[hang,flushmargin]{footmisc}
\algrenewcommand\algorithmicrequire{\textbf{Input:}}
\algrenewcommand\algorithmicensure{\textbf{Output:}}

\makeatletter
\let\NAT@parse\undefined
\makeatother
\usepackage[pagebackref,breaklinks,hidelinks]{hyperref}

\title{\LARGE \bf
Topological Exploration using Segmented Map with \\Keyframe Contribution in Subterranean Environments
}

\author{Boseong Kim, Hyunki Seong, and D. Hyunchul Shim
\thanks{*This research was financially supported by the Institute of Civil Military Technology Cooperation funded by the Defense Acquisition Program Administration and Ministry of Trade, Industry and Energy of Korean government under grant No. UM22206RD2.}
\thanks{*All authors are with the School of Electrical Engineering, Korea Advanced Institute of Science and Technology (KAIST), Yuseong-gu, Daejeon 34141, Republic of Korea (email: \{brian.kim, hynkis, hcshim\}@kaist.ac.kr).}%
}
\begin{document}

\maketitle

\begin{abstract}

Existing exploration algorithms mainly generate frontiers using random sampling or motion primitive methods within a specific sensor range or search space. However, frontiers generated within constrained spaces lead to back-and-forth maneuvers in large-scale environments, thereby diminishing exploration efficiency. To address this issue, we propose a method that utilizes a 3D dense map to generate Segmented Exploration Regions (SERs) and generate frontiers from a global-scale perspective. In particular, this paper presents a novel topological map generation approach that fully utilizes Line-of-Sight (LOS) features of LiDAR sensor points to enhance exploration efficiency inside large-scale subterranean environments. Our topological map contains the contributions of keyframes that generate each SER, enabling rapid exploration through a switch between local path planning and global path planning to each frontier. The proposed method achieved higher explored volume generation than the state-of-the-art algorithm in a large-scale simulation environment and demonstrated a 62\% improvement in explored volume increment performance. For validation, we conducted field tests using UAVs in real subterranean environments, demonstrating the efficiency and speed of our method.

\end{abstract}

\section{INTRODUCTION}
Utilizing unmanned vehicles in subterranean environments for exploration is of paramount importance in the field of robotics, with the objective of reducing human casualties and minimizing property damage. Studies on exploration have significantly evolved over the past five years, greatly influenced by the DARPA Subterranean Challenge\cite{petravcek2021large,shen2012autonomous,kulkarni2022autonomous}. These studies have primarily proposed algorithms based on LiDAR sensors that can operate effectively in visually degraded, large-scale environments with dust or darkness. Among various platforms, UAVs have garnered significant attention due to their high mobility, enabling them to achieve greater exploration efficiency. UAVs possess the advantage of being able to operate independent of terrain, making them suitable for flexible mission execution in environments such as stairs, cliffs, or various irregular terrains\cite{michael2014collaborative,thakur2020nuclear,martz2020survey}, which makes them a preferred platform for exploration. However, despite these advantages, UAVs still face several challenges such as limitations in weight, size, and flight duration. For instance, solving three-dimensional perspective problems within onboard computers with limited computational capabilities and developing efficient exploration algorithms to cover as much area as possible within the given time constraints are demanded.
\begin{figure}[t!]
\begin{center}
\includegraphics[width=1\columnwidth]{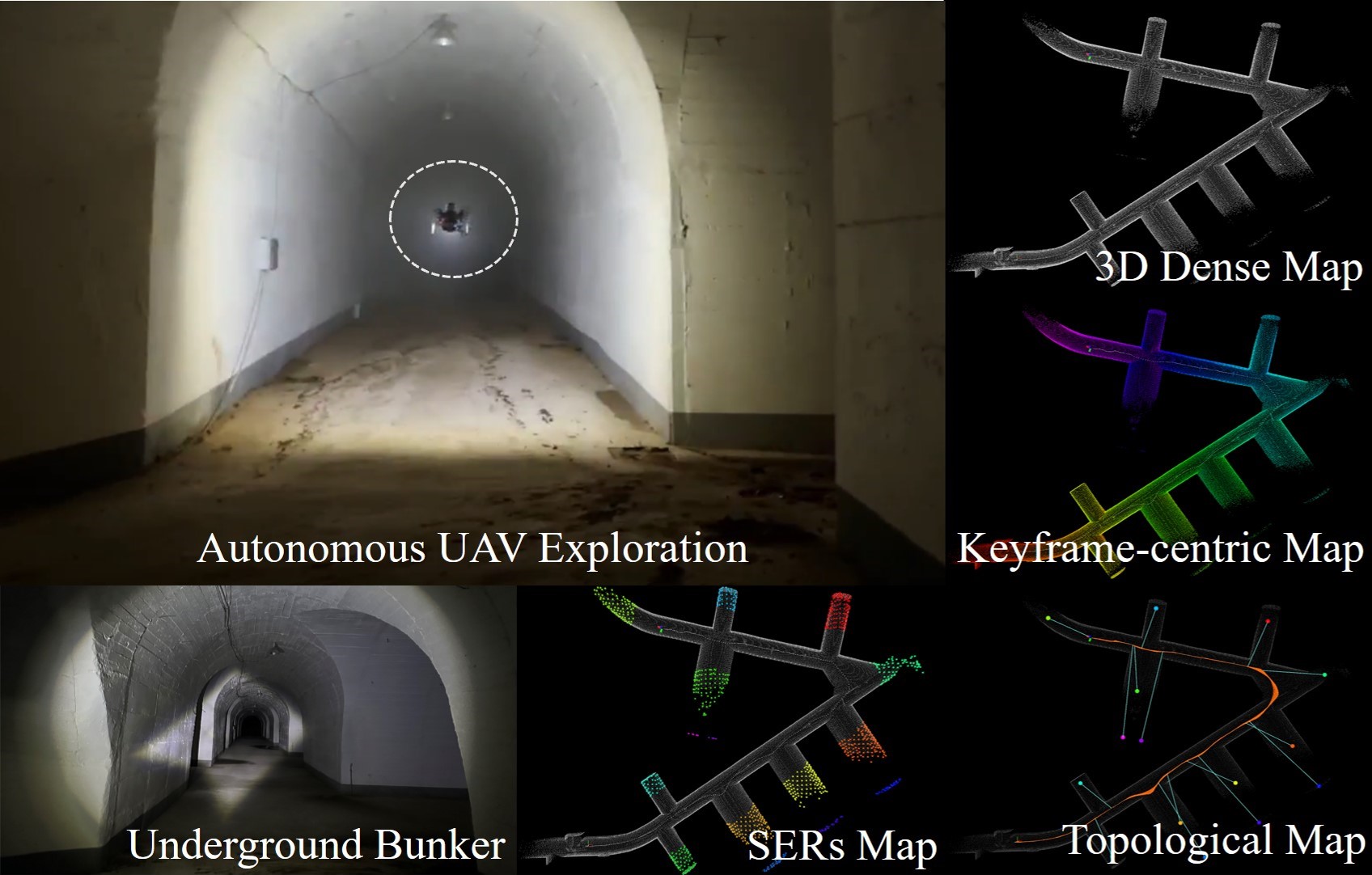}
\end{center}
\caption{An instance of the proposed segmented map-based topological exploration algorithm using a UAV in a subterranean environment.}
\label{Fig:Head}
\end{figure}

To address these issues, we propose a novel topological exploration method, as shown in Fig. \ref{Fig:Head}, that segments the regions from a global-scale perspective and fully leverages the Line-of-Sight (LOS) feature of LiDAR sensors. In more detail, the proposed method utilizes segmented regions and the LiDAR keyframes that have contributed to the generation of these regions to determine the execution of local path planning and global path planning. This reduces unnecessary back-and-forth maneuvers of the UAV during exploration, thereby increasing exploration efficiency. We compared our proposed method with the state-of-the-art algorithm GB planner2 \cite{kulkarni2022autonomous} in a large-scale simulation environment. Furthermore, we demonstrated the high exploration efficiency and speed of the proposed method through field tests using UAVs in real subterranean environments. The primary contributions of this paper are as follows:

\begin{enumerate}
    \item Rapid exploration within large-scale 3D environments through frontier that generated from a global-scale perspective using Segmented Exploration Regions (SERs).
    \item Efficient path planning using a novel topological map that takes into account the relationship between SERs and the most contributed LiDAR keyframes.
    \item Demonstration of the practicality and superiority of the proposed method through large-scale simulation environments and field tests in real subterranean environments.
\end{enumerate}

\section{RELATED WORKS}
Autonomous robotic exploration has been approached in various ways.
A common method involves the use of frontiers\cite{yamauchi1997frontier,cieslewski2017rapid,zhou2021fuel,cao2021tare}, which detect boundaries between known and unknown areas. 
\cite{yamauchi1997frontier} initiated the frontier-based scheme, directing the robot to the nearest frontier in a greedy manner. \cite{cieslewski2017rapid} refined the greedy method, implementing high-speed traversal with minimal velocity changes.
In \cite{zhou2021fuel}, an information structure was presented that includes frontier cells and viewpoints for hierarchical exploration planning.
Similarly, \cite{cao2021tare} introduced a hierarchical framework based on viewpoints and cuboid subspace centroids.

Sampling-based methods explore uncharted spaces by randomly generating viewpoints \cite{bircher2016receding, witting2018history,zhu2021dsvp}, RRT nodes \cite{lindqvist2021exploration,schmid2020efficient,sun2022ada}, or motion primitives\cite{dharmadhikari2020motion}. They are predominantly inspired by the RRT-based algorithms \cite{karaman2011sampling}, emphasizing efficient exploration of complex environments.
\cite{bircher2016receding} is the early work that employs "next best views" (NBVs) to maximize coverage of unknown volumetric spaces.
\cite{witting2018history} enhances the NBV planner to address the curse of dimensionality by reducing sampling space and incorporating a history of explored areas.
In contrast, \cite{dharmadhikari2020motion} generates a sequence of motion primitives via randomized control space sampling.

Recent research has increasingly focused on large-scale subterranean environments. Notably, the field of underground exploration has seen a surge in interest due to the DARPA Subterranean Challenge\cite{darpasubt}. To explore large-scale, multi-branched spaces, topology\cite{silver2006topological,yang2021graph} and graph-based approaches\cite{dang2019graph, kulkarni2022autonomous} have been proposed for representing exploration regions. In a recent effort, \cite{yang2021graph} employs convex polyhedra to separate 3D regions, aiding the selection of local exploration directions. This approach leverages the separated regions to generate frontier points and select local exploration directions. The graph-based planners\cite{dang2019graph}, on the other hand, utilize a rapidly-exploring random graph to identify collision-free local paths that maximize the exploration gain. These methodologies are further enhanced by incorporating global path planning, which assists homing operations \cite{dang2019graph, petravcek2021large} and re-positioning \cite{yang2021graph,kulkarni2022autonomous} of exploring robots.
While the results are promising, these random sampling-based methods generate a redundant graph representation in a local 3D space and require intensive computation, affecting the efficiency of exploration planning.

In this study, we introduce a novel frontier generation scheme designed from a global-scale perspective to minimize redundancies, such as back-and-forth maneuvers, thereby enhancing the exploration performance. We also present an exploration strategy that employs the keyframe contribution to facilitate efficient and rapid exploration in large-scale environments.

\section{Problem Statement}
When launching robots for exploration in unknown areas, LiDAR-based localization (LiDAR Inertial Odometry or SLAM) is imperative. The entirety of map points $V\subset\mathbb{R}^3$, generated from LiDAR-based localization, can be partitioned into the explored region $V_\text{cover}$ and the unexplored region $V_\text{uncover}$, utilizing the keyframes $K\subset\mathbb{R}^3$ representing the map-generating positions and the coverage $\zeta_{coverage}$ ($V = V_\text{cover} \cup V_\text{uncover}$). When the map points generated from $i$-th keyframe $K_i\in\mathbb{R}^3$ are $Z_i\subset\mathbb{R}^3$, then $V$ can be expressed as $V=\{Z_{i,i\in 0,1,2,\cdots,k}\}$, and $V_\text{cover}$ is consists of all \{x,y,z\} points in $V$ that have a Euclidean distance from each element of $K$ within $\zeta_{coverage}$. The primary objective of the proposed exploration algorithm is to reduce the $V_\text{uncover}$, ultimately satisfying $V = V_\text{cover}$.
\section{Proposed Approach}
\subsection{Segmented Exploration Regions (SERs)}\label{SER}
\begin{figure}[t!]
\begin{center}
\includegraphics[width=1\columnwidth]{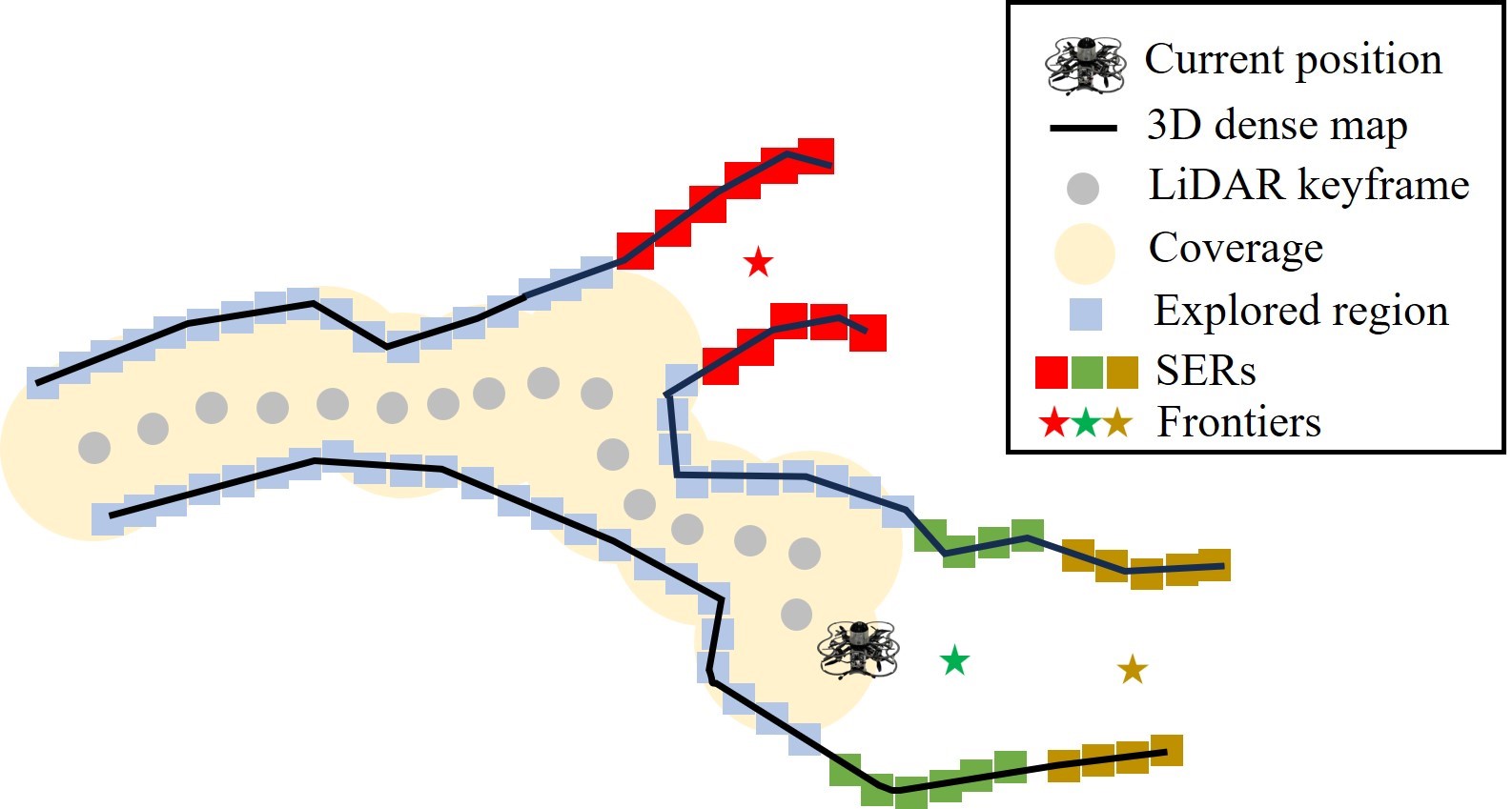}
\end{center}
\caption{An overview of Segmented Exploration Regions (SERs) generation. The black lines represent 3D map points $V$ generated from LiDAR keyframes $K$, and the blue squares denote explored regions $V_\text{cover}$ included in coverage $\zeta_{coverage}$. SERs are generated by applying Euclidean distance clustering to unexplored regions $V_\text{uncover}$, with squares of the same color indicating the same SER. Frontier corresponding to each SER is represented by star shape with the same color.}
\label{Fig:SERs}
\end{figure}
\begin{algorithm}[t]    
    \caption{Real-time Generation of SERs and Frontiers}\label{alg1}
    \begin{algorithmic}[1]
        \Require{All keyframes $K$, all map points $V$, voxel size $v_\text{down}$}
        \Ensure{Segmented Exploration Regions (SERs) $V_\text{SERs}$}
        \If{$\text{New keyframe $K_t$ is generated}$}
            \State $K$.Pushback($K_t$), $V$.Pushback($Z_t$) 
            \State $V_\text{uncover}$, $V_\text{cover}$ $\gets$ $\emptyset$ \Comment{Initialization}
            \State ${}^{\prime}V$ $\gets$ $\text{Voxelization}(V)$ $\text{with}$ $v_\text{down}$ \Comment{Down-sampling}
            \For{All $\{x,y,z\}$ points in ${}^{\prime}V$}
            \State Cover = $false$ 
            \For{All $K_i$ in $K$}
            \If{$\Vert K_i-\{x,y,z\} \Vert\leq \zeta_{\text{coverage}}$}
            \State Cover = $true$ \Comment{Within coverage}
            \State $V_\text{cover}$.\textbf{Pushback}($\{x,y,z\}$)
            \State \textbf{Break}
            \EndIf
            \EndFor
            \If{\text{Cover}$=false$}\Comment{Unexplored points}
            \State $V_\text{uncover}$.\textbf{Pushback}($\{x,y,z\}$)
            \EndIf
            \EndFor
        \State $V_\text{SERs}\gets\textbf{EuclideanDistanceClustering}(V_\text{uncover})$
        \For{All $V^{j}_\text{SER}$ in $V_\text{SERs}$}
        \State $g^j\gets\textbf{Centroid} (V^{j}_\text{SER})$ \Comment{Frontier generation}
        \EndFor
        \EndIf
    \end{algorithmic}
\end{algorithm}
For efficient exploration in large-scale underground environments, we divide $V_\text{uncover}$ into multiple Segmented Exploration Regions (SERs) $V_\text{SERs}\subset V_\text{uncover}$, using Euclidean distance clustering techniques as shown in Fig. \ref{Fig:SERs}. Unlike existing methods that generate frontiers within a specific sensor range, the proposed exploration method divides the three-dimensional space into segments based on the geometric characteristics of the $V_\text{uncover}$, generating frontiers at a global scale. $V_\text{SERs}=V^{0}_\text{SER} \cup V^{1}_\text{SER} \cup V^{2}_\text{SER} \cdots V^{k}_\text{SER}$ are utilized for the frontier generation, and the centroid of each $V^{j}_\text{SER}\subset\mathbb{R}^3$ is considered as the frontier $g^j\in\mathbb{R}^3$ that should be reached to cover the $V^{j}_\text{SER}$ for exploration. As exploration progresses and $K$, $V$, $V_\text{cover}$, and $V_\text{uncover}$ are updated, the $V_\text{SERs}$ is also updated in real-time. 

As described in Algorithm \ref{alg1}, the generation of SERs is carried out when a map is updated (when a new keyframe is generated) in LiDAR-based localization. The updated map is down-sampled for computational efficiency and is divided into $V_\text{cover}$ and $V_\text{uncover}$ by comparing it with the keyframes generated so far. For $V_\text{uncover}$, we apply Euclidean distance clustering techniques to generate $V_\text{SERs}$. Finally, for each $V^{j}_\text{SER}$, the corresponding frontier point $g^j$ is generated, which is considered as a candidate for the robot to reach in order to reduce $V_\text{uncover}$. The reason keyframe generation serves as the criterion for SERs generation lies in the need to systematically manage all generated map points during exploration, pairing them with corresponding $K_i$ and $Z_i$. 
Given the inherent nature of LiDAR sensors, $Z_i$ maintain a Line-of-Sight (LOS) trajectory from $K_i$, a pivotal element for the generation of frontier edge, which we aim to explain in Section \ref{Frontier}.

\subsection{Real-time Graph Generation with LiDAR Keyframes}\label{Graph}
When a new keyframe is generated, we generate edges that account for connectivity between each pair of nodes (keyframes) using $K$ and $V$. The graph $G$, composed of nodes and edges, serves as the foundation for the robot's global path planning within the large-scale environment and is updated in real-time during exploration. The edge between the $i$-th node $K_i$ and the $j$-th node $K_j$ is determined by performing collision checks with the sub-map $V_s \subset V$. $V_s$ is represented by a set of $Z$ extracted from $k$ keyframes $K^k =\{K_{k_0}, K_{k_1}, \cdots, K_{k_{k-1}}\}$ using the K-nearest neighbors (KNN) algorithm.

Similar to the $V_\text{SERs}$ generation, the generation of the graph $G$ is performed whenever a new keyframe is generated. The proposed graph generation method not only considers connectivity between $K_{t-1}$ and $K_t$ but also involves examining the connectivity with the $k$ nearest nodes. This approach is essential for efficient global path planning in complex environments with multiple branches, characteristic of large-scale scenarios. When a new keyframe $K_t$ is generated, it is added to the keyframe array $K$, and its corresponding $Z_t$ is added to the map point array $V$. Subsequently, using the KNN algorithm, the indices of $k$ nearest keyframes are extracted, and a sub-map $V_s$ for collision checking is swiftly generated by leveraging the features of paired $K$ and $V$. 

Collision checking between two nodes and sub-map points is conducted along line segments and sub-map points. If the vectors from each node to sub-map points form an acute angle with the line segment connecting the nodes, collision checking is performed using the Euclidean distance between the point and the line segment. Otherwise, collision checking is carried out using the Euclidean distance between the point and the node forming an obtuse angle.

\begin{algorithm}[t]    
    \caption{Frontier Graph Generation with Contribution}\label{alg3}
    \begin{algorithmic}[1]
        \Require{All keyframes $K$, all map points $V$}
        \Ensure{Frontier graph $G_\text{F}$, keyframe-centric map $V_\text{key}$}
        \If{$\text{New keyframe $K_t$ is generated}$}
            \State $G_\text{F}$, $V_\text{uncover}$, $V_\text{cover}$ $\gets$ $\emptyset$ \Comment{Initialization}
            \State $K$.\textbf{Pushback}($K_t$), $V$.\textbf{Pushback}($Z_t$)
            \State $V_\text{key}$.\textbf{Pushback}(\{$Z_t, K_t$\})\Comment{Keyframe-centric map}
            \State $g$, $V_\text{SERs}\gets\textbf{GenerateSERs}(K,V,v_\text{down})$ \Comment{Alg. \ref{alg1}}
            \For{All $V^{j}_{\text{SER}}$ in $V_\text{SERs}$}
            \State $K^j_\text{SER}$ $\gets$ $\emptyset$
            \For{All $\{x,y,z\}$ points in $V^{j}_{\text{SER}}$}
            \State $K_{E}$ $\gets$ \textbf{ExtractKeyframe}$(\{x,y,z\},V_\text{key})$
            \State $K^{j}_{\text{SER}}$.\textbf{Pushback}$(K_E)$
            \EndFor
            \State $K_\text{Highest}$ $\gets$ \textbf{HighestContribution}$(K^{j}_{\text{SER}})$
            \State $G_{\text{F}}$.\textbf{addNode}($g^j$) 
            \State $G_{\text{F}}$.\textbf{addEdge}($g^j$,$K_\text{Highest}$) 
            \EndFor
        \EndIf
    \end{algorithmic}
\end{algorithm}
\subsection{Frontier Graph Generation with Keyframe Contribution}\label{Frontier}
In this section, we present a method for generating a graph between frontiers $g=\{g^{j,j\in0,1,\cdots,l}\}$ and specific node $K_i$ based on the contribution of keyframes. In Section \ref{SER}, we explained how $V_\text{SERs}$ are generated for frontiers. These frontiers are generated from $V_\text{SERs}$ by applying Euclidean distance clustering to $V_\text{uncover}$, making them a result of a global-scale perspective.

If we know which keyframes contributed to the points composing each $V^{j}_{\text{SER}}$, we can extract the keyframe that had the most significant contribution to a particular frontier. As previously mentioned, our exploration method is based on the LOS characteristic of LiDAR sensor points. $V^j_\text{SER}$ are constituted by points generated from multiple keyframes $K^j_\text{SER}\subset K$ but are generated based on geometric features of 3D dense map. Therefore, if a specific keyframe most contributed to $V^{j}_{\text{SER}}$, we consider LOS to that frontier $g^j$, even if not all points in that $V^{j}_{\text{SER}}$ were generated by the same keyframe.
\begin{figure}[t!]
\begin{center}
\includegraphics[width=1\columnwidth]{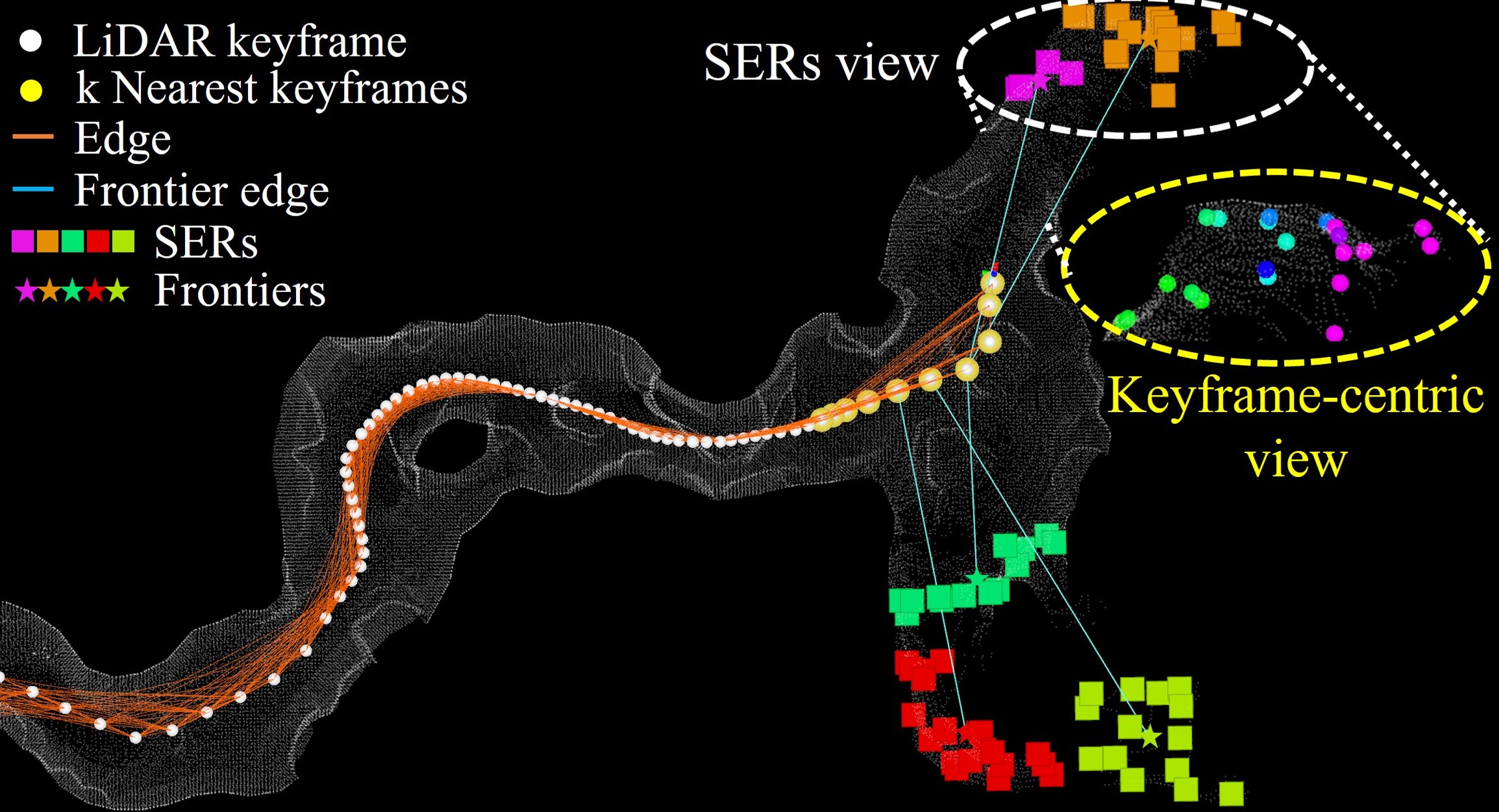}
\end{center}
\caption{Comparison between the SERs map $V_\text{SERs}$ and the keyframe-centric map $V_\text{key}$. Each SER visible in the SERs view is generated based on the geometric features of the $V_\text{uncover}$. The points visible in the keyframe-centric view represent the keyframes (colored) that generated those points. The frontier graph $G_\text{F}$ is generated using the keyframe $K_\text{Highest}$ that made the most significant contribution to the generation of each SER and its corresponding frontier.}
\label{Fig:keyframe-centric}
\end{figure}
Because we manage $K$ and $V$ as pairs, we introduce a novel map representation named the keyframe-centric map $V_\text{key}=\{V^{i}_{\text{key}}\}$. $V_\text{key}\subset\mathbb{R}^6$ includes each $Z_i$ that constitutes $V$ along with its corresponding $K_i$. The $j$-th point within the map points generated from the $i$-th keyframe can be denoted as $Z_{i,j}\in\mathbb{R}^3$, and from the perspective of $V_\text{key}$, the corresponding point can be represented as $V^{i,j}_\text{key}=\{Z_{i,j},K_i\}$. The generation of frontier edges based on keyframe contributions is detailed in Algorithm \ref{alg3}.

The generation of the frontier graph $G_\text{F}$ begins by parallelly updating the keyframe-centric map $V_\text{key}$ with the newly generated $Z_t$ and $K_t$. Following this, Algorithm \ref{alg1} in Section \ref{SER} is applied to generate $g$ and $V_\text{SERs}$. For each $V^{j}_{\text{SER}}$ comprising $V_\text{SERs}$, the keyframe that contributes the most to the generation of that $V^{j}_{\text{SER}}$ is extracted. As shown in Fig. \ref{Fig:keyframe-centric}, for all $\{x,y,z\}$ points forming $V^{j}_{\text{SER}}$, we concurrently search for and extract the corresponding $V^{i,j}_\text{key}$ from the parallelly generated $V_\text{key}$, along with the keyframe information, generating all keyframes $K^j_\text{SER}$ that generated  $V^{j}_{\text{SER}}$. Finally, the keyframe $K_\text{Highest}$ with the highest contribution within $K^j_\text{SER}$ is extracted to construct the graph between the frontier $g^j$ generated from $V^{j}_{\text{SER}}$ and $K_\text{Highest}$. It's important to note that $G$ described in Section. \ref{Graph} is utilized for the robot's global path planning, necessitating collision checks. However, $G_\text{F}$ is employed to just specify frontiers under the assumption of LOS from specific nodes, and thus, collision checks are not performed.
\subsection{Global Path Planning within Topological Map}\label{Topological}
In Sections \ref{Graph} and \ref{Frontier}, we described the generation process of $G$ and $G_\text{F}$. In this section, we aim to describe efficient global path planning for exploration in large-scale 3D environments characterized by multiple branches, utilizing the topological map composed of $G$ and $G_\text{F}$. When a new keyframe $K_t$ is generated, $V_\text{SERs}$, $g$, $G$, and $G_\text{F}$ are updated. First, for each frontier $g^j$ generated from each $V^{j}_{\text{SER}}$, we use the proposed exploration score to identify the frontier $g^{j_\text{best}}$ with the highest score. We then prioritize the exploration of $g^{j_\text{best}}$. The proposed exploration score can be expressed as 
\begin{align}
    \textbf{Expl}&\textbf{orationScore}(g^j)=\nonumber\\ 
    &\frac{w_\textit{Vol}\textbf{\textit{Volume}}(V^{j}_{\text{SER}})}{ w_\textit{Dir}\textbf{\textit{Direction}}(\psi_t, \psi_{g^j})\cdot w_\textit{Dis}\textbf{\textit{Distacne}}(K_t, g^j)}.\label{eqA} 
\end{align}
\begin{figure}[t!]
\begin{center}
\includegraphics[width=1\columnwidth]{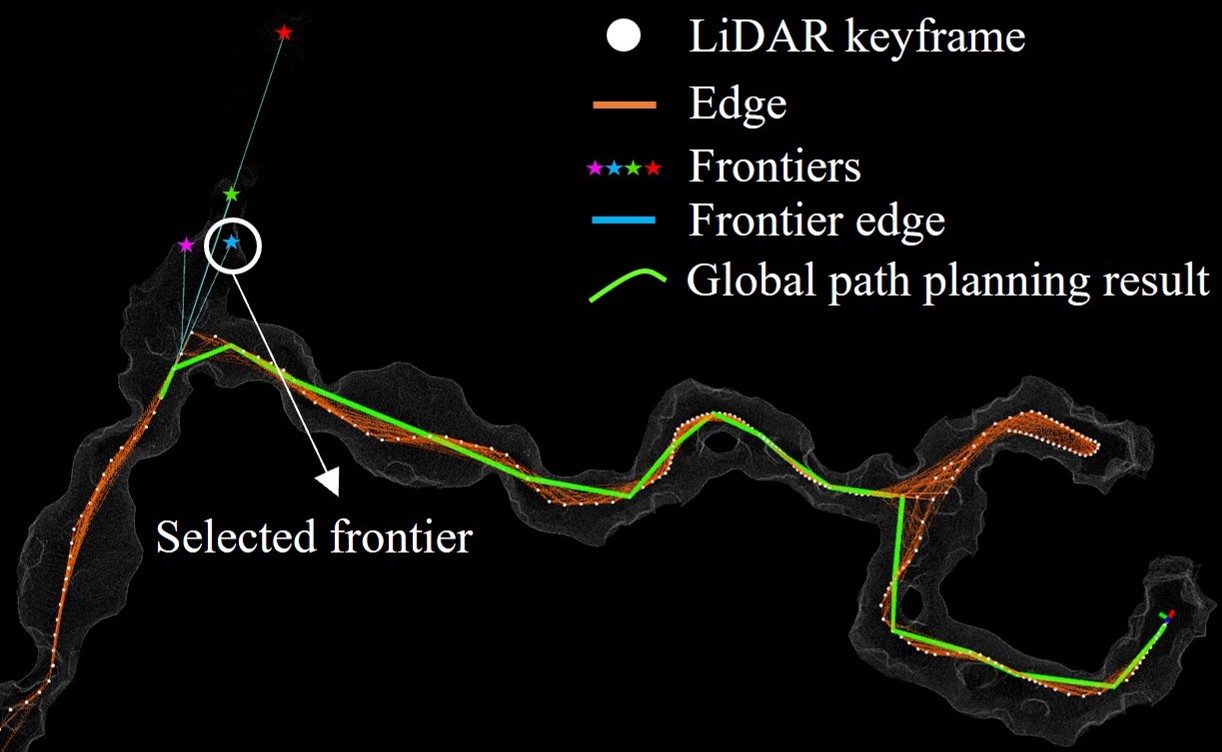}
\end{center}
\caption{Global path planning results within the proposed topological map. If the SER that generated the selected frontier does not include any points of $Z_t$ generated by the current keyframe $K_t$, global path planning is performed using $G$ and $G_\text{F}$.}
\label{Fig:Globalpath}
\end{figure}
\begin{figure}[t]
\begin{subfigure}[b]{1\columnwidth}
\includegraphics[width=1\columnwidth]{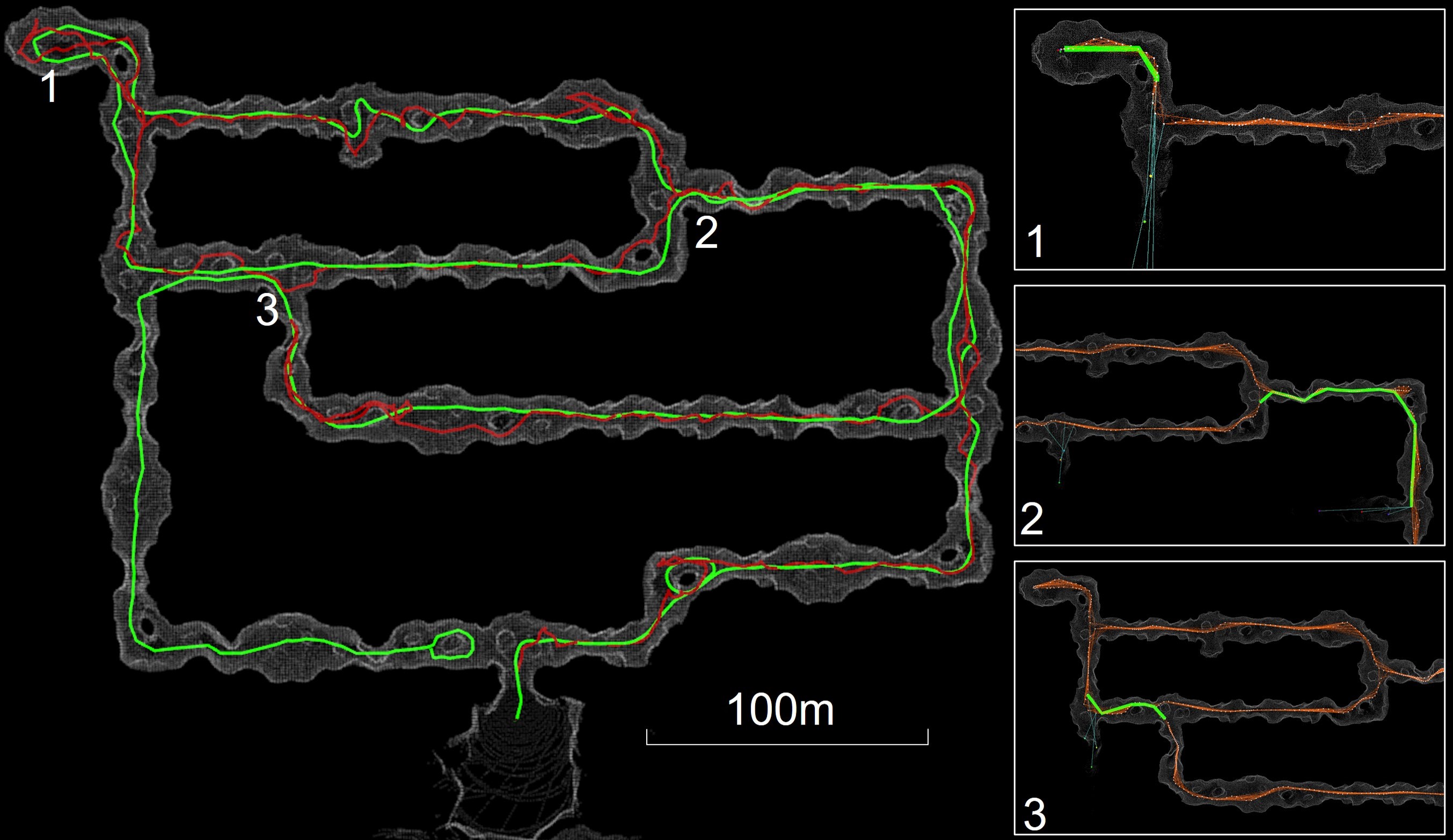}
\caption{}
\label{Fig:cave02}
\end{subfigure}
\begin{subfigure}[b]{1\columnwidth}
\includegraphics[width=1\columnwidth]{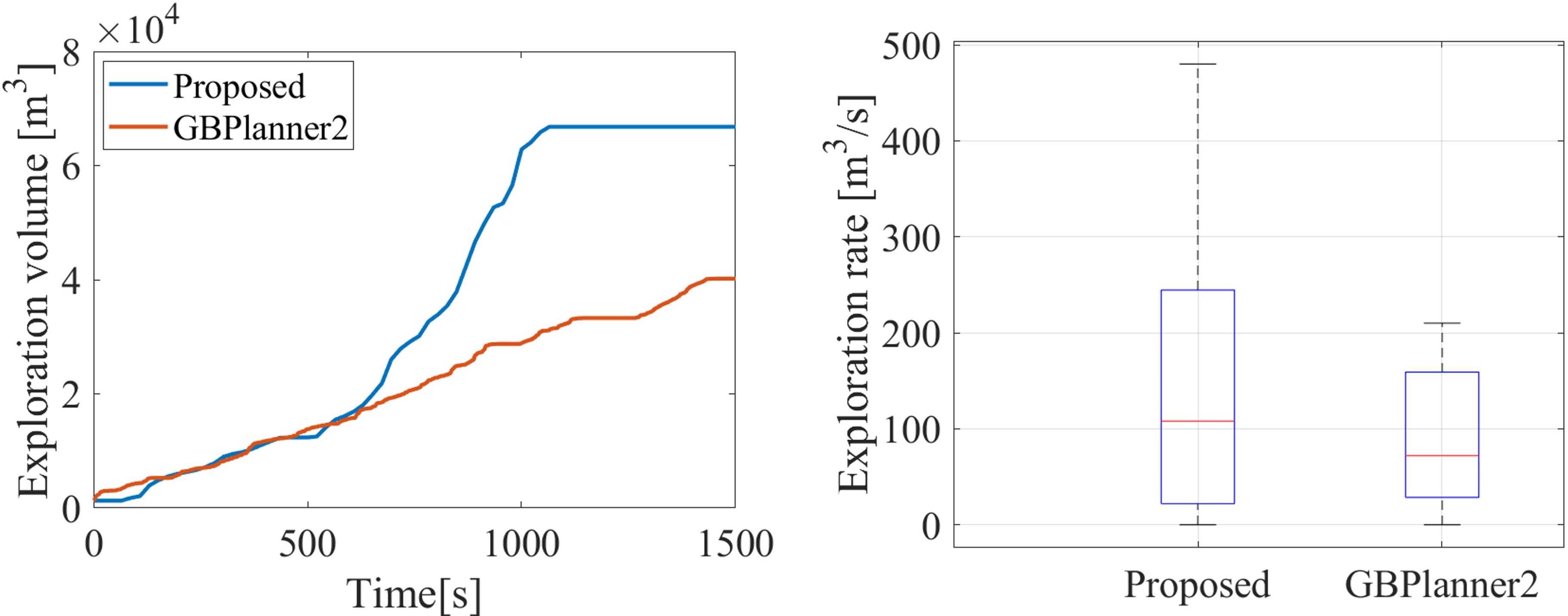}
\caption{}
\label{Fig:caveperform}
\end{subfigure}
\caption{Comparison of exploration performance between the proposed method and GB planner2 using UAV. (a) Comparison of exploration trajectory between the proposed method (green) and GB planner2 (red). The inset figure on the right illustrate the results of global path planning triggered within the topological map through our exploration planning strategy, shown as the green path. (b) Quantitative comparison of explored volume over time and the explored volume increment.}
\label{Fig:Cave}
\end{figure}
\begin{algorithm}[t!]    
    \caption{Global Path Planning within Topological Map}\label{alg4}
    \begin{algorithmic}[1]
        \Require{Graph $G$, frontier graph $G_\text{F}$}
        \Ensure{Path $\Sigma(K_t, K_\text{Highest})$}
        \If{$\text{New keyframe $K_t$ is generated}$}
        \State $g$, $V_\text{SERs}\gets\textbf{GenerateSERs}(K,V,v_\text{down})$ \Comment{Alg. \ref{alg1}}
        \State $G$$\gets$$\textbf{GenerateGraph}(K,V)$ \Comment{Sec. \ref{Graph}}
        \State $G_\text{F}$$\gets$$\textbf{GenerateFroniterGraph}(K,V)$ \Comment{Alg. \ref{alg3}}
        \If{$\text{Robot reaches the $g^{j_{\text{best}}}$}$}
            \State $g^{j_{\text{best}}}$ $\gets$ $\textbf{ExplorationScore}(g)$ \Comment{\eqref{eqA}}
            \If{$\text{$V^{j_\text{best}}_{\text{SER}}$ contains any point from the $Z_t$}$}
            \State $\textbf{\textit{Local path planning towards the}}$ $g^{j_{\text{best}}}$
            \Else \Comment{Global path planning}
            \State $K_\text{Highest}$ $\gets$ $G_\text{F}.\textbf{findEdge}(g^{j_{\text{best}}})$
            \State $\Sigma(K_t, K_\text{Highest})$ $\gets$ $\textbf{Dijkstra}(G, K_t, K_\text{Highest})$
            \State $\textbf{\textit{Path following along the }}{\Sigma(K_t, K_\text{Highest})}$
            \EndIf
        \EndIf
        \EndIf
    \end{algorithmic}
\end{algorithm}

As shown in \eqref{eqA}, the proposed exploration score is composed of distance, volume, and direction factors. $\textbf{\textit{Distacne}}(K_t, g^j)$ represents the distance between the current node $K_t$ and frontier $g^j$, with a higher score computed for closer frontiers, emphasizing proximity. Secondly, $\textbf{\textit{Volume}}(V^{j}_{\text{SER}})$ represents the number of points constituting $V^{j}_{\text{SER}}$, and a higher score is computed for $g^j$ generated with a greater number of points in $V^{j}_{\text{SER}}$. This design aims to efficiently map as large volume as possible within a limited time. Lastly, $\textbf{\textit{Direction}}(\psi_t, \psi_{g^j})$ represents the difference between the current robot's yaw and the direction towards $g^j$. A higher score is computed for frontiers with minimal direction difference, aiming to reduce back-and-forth maneuvers during exploration and enhance exploration efficiency. Note that, $w_\textit{Dis}$, $w_\textit{Vol}$, and $w_\textit{Dir}$ represent the weights for the distance, volume, and direction factors, respectively.

Using the proposed exploration score, once the frontier $g^{j_{\text{best}}}$ with the highest score is selected, we proceed to decide whether to perform global path planning or local path planning towards $g^{j_{\text{best}}}$. Fortunately, the proposed exploration algorithm leverages the keyframe-centric map $V_{\text{key}}$ to determine which keyframes contributed to generating the selected $V^{j_\text{best}}_{\text{SER}}$, allowing us to ascertain whether $V^{j_\text{best}}_{\text{SER}}$ contains any point from the $Z_t$ generated from the $K_t$. If it does, considering the LOS from the $K_t$ to $V^{j_\text{best}}_{\text{SER}}$, exploration towards  $g^{j_{\text{best}}}$ is conducted using local path planning. On the other hand, if $V^{j_\text{best}}_{\text{SER}}$ does not contain any point from the $Z_t$, Non-Line of Sight (NLOS) is considered, leading to global path planning towards the $g^{j_{\text{best}}}$ within the topological map composed of $G$ and $G_\text{F}$. Global path planning using the topological map is detailed in Algorithm \ref{alg4}, and the result is shown in Fig. \ref{Fig:Globalpath}.

\begin{figure}[t!]
\begin{center}
\includegraphics[width=1\columnwidth]{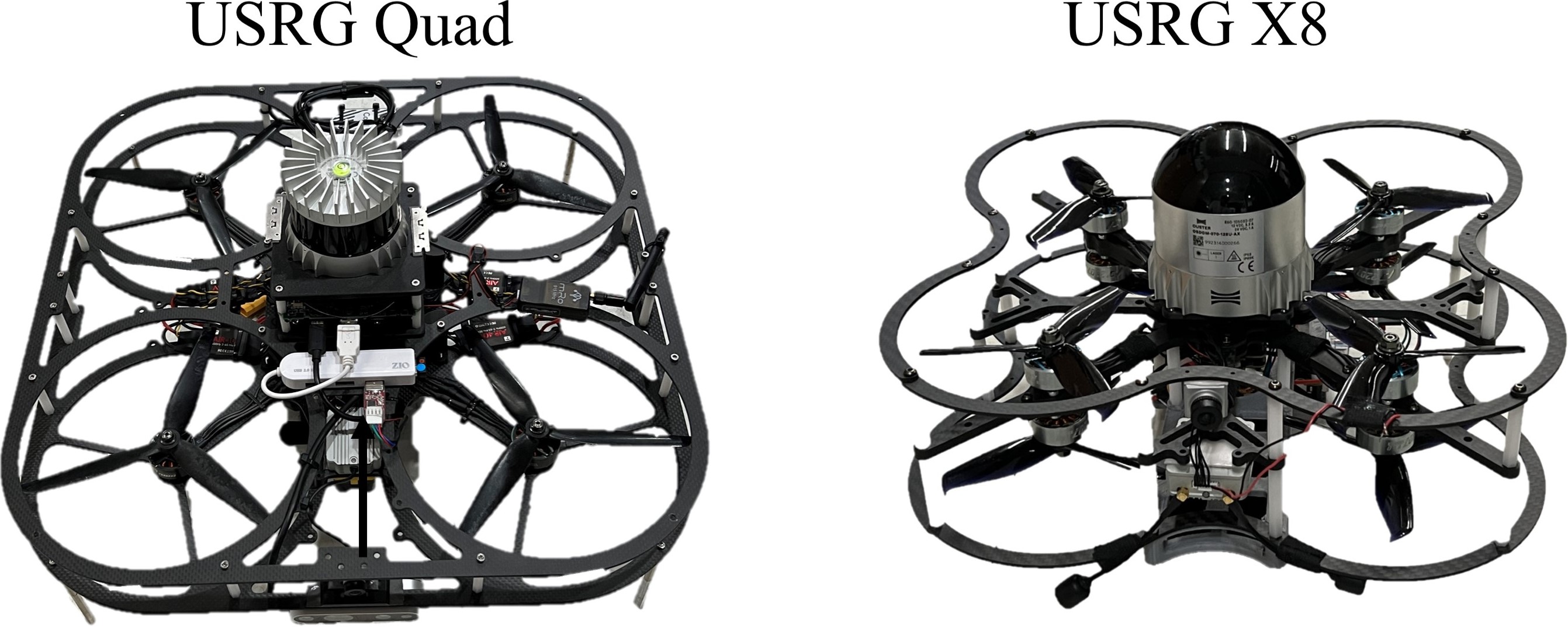}
\end{center}
\caption{The two custom aerial platforms used for field testing, the USRG Quad and the USRG X8.}
\label{Fig:Drones}
\end{figure}
\section{Experiments}
\begin{figure*}[t]
\begin{center}
\includegraphics[width=1.0\textwidth]{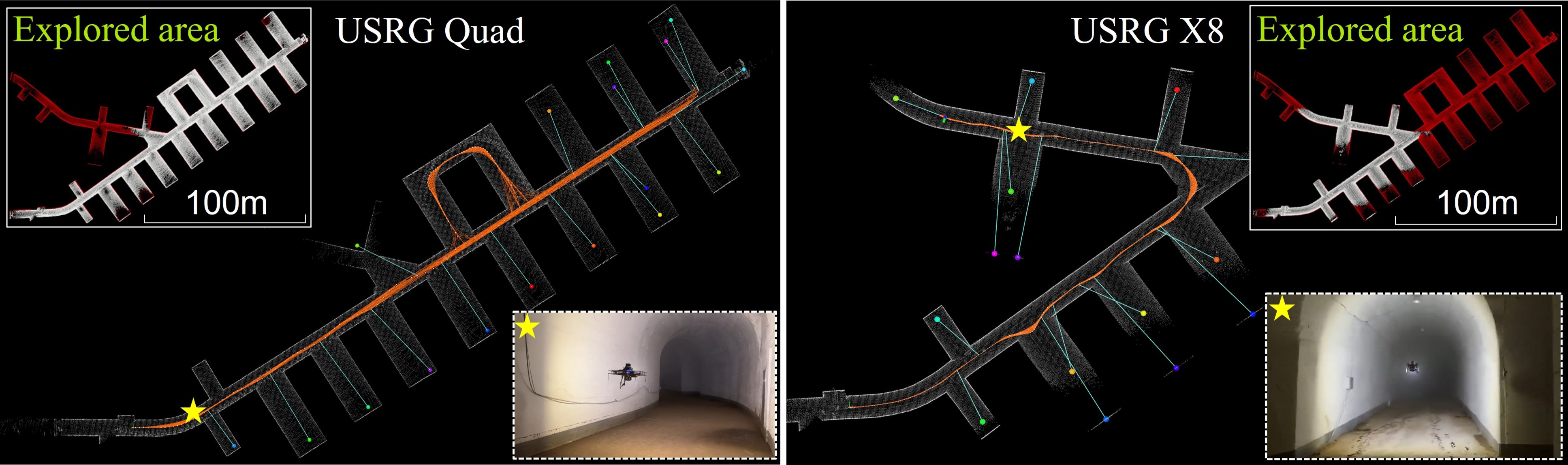}
\end{center}
\caption{Exploration results using USRG Quad and USRG X8 in the subterranean environment located in South Korea. The white map, orange lines, light blue lines, and colored points represent the 3D dense map, graph edges, frontier graph edges, and frontiers generated during the exploration, respectively.}
\label{Fig:field}
\end{figure*}
\begin{figure}[t]
\begin{center}
\includegraphics[width=1\columnwidth]{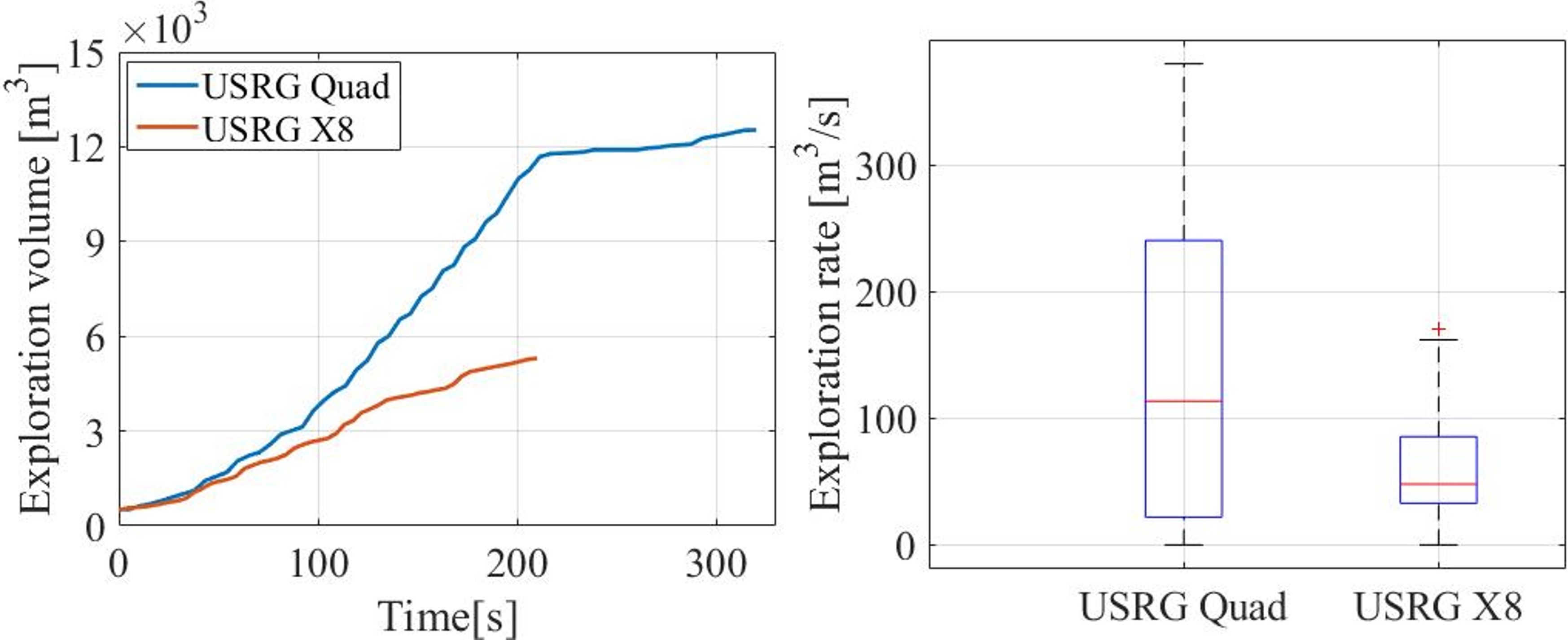}
\end{center}
\caption{Comparing the quantitative exploration performance of two UAVs equipped with different LiDAR sensors. The USRG Quad equipped with a LiDAR sensor with a longer maximum range, demonstrates higher exploration efficiency compared to the USRG X8.}
\label{Fig:USRG_performance}
\vspace{-1em}
\end{figure}
\subsection{Simulation Based Evaluation}
In this section, the performance of our method was compared with the state-of-the-art GB planner2 \cite{kulkarni2022autonomous} algorithm in the 3D cave simulation environment provided by the DARPA Subterranean virtual competition using UAV. To evaluate the performance, two factors, explored volume increment per second $(m^3/s)$ and explored volume over time $(m^3)$, were compared, and the results are shown in Fig. \ref{Fig:Cave}.

In the simulation environment, both algorithms used a LiDAR sensor with a Horizontal Field of View (HFOV) of 360$^\circ$, a Vertical Field of View (VFOV) of 45$^\circ$, a maximum sensor range of 80$m$ with 32 channels. In the simulation environment, the proposed method had a coverage $\zeta_{coverage}$ of 15$m$, a down-sampling parameter $v_{\text{down}}$ of 5$m$, and $k$ set to 10. A motion primitive-based planning algorithm was used for our local path planning, and both exploration algorithms had a maximum flight speed of 2$m/s$. As shown in Fig. \ref{Fig:Cave}, the proposed method exhibits fewer back-and-forth maneuvers based on our novel planning strategy (detailed in Section. \ref{Topological}) and outperformed the GB planner2 in both factors. Also, the proposed method exhibits smoother exploration paths compared to GB planner2, achieved through frontier generation from a global-scale perspective and topological exploration based on keyframe contributions. The proposed method generates a larger volume within the same time compared to GB planner2, demonstrating a 62\% improvement in median value of the volume increment.
\subsection{Experimental Evaluation}
For a more comprehensive analysis of the proposed method, we conducted field tests using UAVs in a real subterranean environment. The performance of the proposed method varies depending on the specification of LiDAR sensors, as it relies on the geometric features of the 3D dense map generated during the exploration. Therefore, we utilized two aerial platforms equipped with different LiDAR sensors to perform separate explorations within the same environment and subsequently analyzed their performance.
\subsubsection{Hardware and Software Setup}
Two aerial platforms, as shown in Fig. \ref{Fig:Drones}, the USRG Quad and the USRG X8, were used. The USRG Quad is an 8-inch quadcopter platform equipped with an Ouster OS1 32-channel LiDAR with an HFOV of 360$^\circ$, VFOV of 45$^\circ$, and an effective range of 90$m$. The USRG X8 is a 5-inch coaxial octocopter platform equipped with an Ouster OS Dome 128-channel LiDAR with an HFOV of 360$^\circ$, VFOV of 180$^\circ$, and an effective range of 20$m$. Both platforms used an Intel NUC computer with a 6-core i7-10710U CPU for real-time onboard computation, and motion primitive-based planning algorithms were employed for local path planning. Localization was achieved using our previous work\cite{kim2023adaptive}, LiDAR Inertial Odometry, which provides keyframes, corresponding LiDAR scans, and UAV positions. During the field tests, the maximum flight speed was set to 1.5$m/s$ for the USRG Quad, and 1.2$m/s$ for the USRG X8, and both platforms were equipped with LED for providing visual information in dark environments.
\subsubsection{Segmented map-based Topological Exploration}
The proposed method for the field test was set up with both UAVs having a coverage $\zeta_{coverage}$ of 7$m$, a down-sampling parameter of 2$m$, and $k$ set to 10. Both UAVs started exploration from the entrance of an underground bunker and performed exploration until their batteries were exhausted. The exploration results for USRG Quad and USRG X8 are shown in Fig. \ref{Fig:field}, while the exploration performance is shown in Fig. \ref{Fig:USRG_performance}. USRG Quad covered approximately 354$m$ by flying for 320$s$ at an average speed of 1.1$m/s$, while USRG X8 covered approximately 168$m$ by flying for 210$s$ at an average speed of 0.8$m/s$. The inset figures labeled as 'Explored area' in Fig. \ref{Fig:field} represent the maps (white map) generated by the two UAVs inside the underground bunker (red map). USRG Quad covered approximately 79\% of the entire map, while USRG X8 covered approximately 43\%. Additionally, the inset figures marked with a star shape show the flight view of the two UAVs from corresponding positions. The quantitative exploration performance based on the maximum range of the LiDAR sensor is shown in Fig. \ref{Fig:USRG_performance}. As shown in Fig. \ref{Fig:USRG_performance}, on a median value basis, USRG Quad had a volume increment of 118$m^3/s$, while the USRG X8 had a 51$m^3/s$. Through the field test using two UAVs, the proposed method has demonstrated the ability to achieve fast and efficient exploration in a real subterranean environment by leveraging our novel keyframe contribution-based topological map and exploration planning strategy.
\section{Conclusions}
In this paper, we proposed a topological exploration algorithm using keyframe contribution. Unlike existing methods that generate frontiers within a specific sensor range or search space, our method generates frontiers from a global-scale perspective, enhancing exploration efficiency in large-scale environments. Using the keyframe-centric map and SERs map, we generate a frontier graph by considering which keyframe contributes the most when frontiers are generated. Finally, by using data structures pairing keyframes with their corresponding scans, we design planning strategies for exploring specific frontiers, enabling rapid exploration. The proposed method surpasses the state-of-the-art GB planner2 algorithm in a large-scale cave simulation environment, showcasing a 62\% improvement in the median value of volume increment. Moreover, it has demonstrated rapid and efficient exploration performance in real subterranean environments through field tests. In the future, our goal is to extend the proposed method to multi-robot or heterogeneous robot applications in various unstructured environments.

\clearpage
\bibliographystyle{unsrt}
\bibliography{reference}

\end{document}